\documentclass[runningheads]{llncs}
\usepackage[T1]{fontenc}
\usepackage{graphicx}
\usepackage{booktabs}
\usepackage{amsmath}
\usepackage{amssymb}
\usepackage{tabularx}
\usepackage{xcolor}
\usepackage{listings}
\usepackage{multirow}

\lstdefinestyle{promptstyle}{
    basicstyle=\ttfamily\scriptsize,
    breaklines=true,
    frame=single,
    backgroundcolor=\color{gray!8},
    rulecolor=\color{gray!45},
    xleftmargin=0.5em,
    xrightmargin=0.5em,
    framexleftmargin=0.4em,
    framexrightmargin=0.4em,
    aboveskip=0.5em,
    belowskip=0.5em,
    columns=fullflexible,
    keepspaces=true
}

\begin{document}
\title{Beyond Vision-Language Models: Visual-Semantic Affect Graphs for Emotion Recognition in Artworks}
\titlerunning{Visual-Semantic Affect Graphs for Emotion Recognition in Artworks}
%

\author{
Ngoc-Duy Mai\inst{1,2,3}\orcidID{0009-0008-8525-6238}\thanks{Corresponding author.} \and
Ha-Thanh Nguyen\inst{4}\orcidID{0000-0003-2794-7010} \and
Hideaki Takeda\inst{1,2}\orcidID{0000-0002-2909-7163}
}

\authorrunning{Duy et al.}

\institute{
National Institute of Informatics, Tokyo, Japan \and
The Graduate University for Advanced Studies, SOKENDAI, Kanagawa, Japan \and
Research and Development Center for Large Language Models, NII, Tokyo, Japan \and
College of Engineering \& Computer Science, VinUniversity, Hanoi, Vietnam \\
\email{duymn@nii.ac.jp}
}

%

%
\maketitle              

\vspace{-10pt}

\begin{abstract}
Emotion recognition in artworks remains challenging because affective meaning is often subjective, implicit, and shaped by relationships between visual form and semantic associations. While vision-language models provide strong multimodal representations, direct image-to-emotion prediction may overlook explicit relations among formal, semantic, and affective cues that support emotional interpretation. This paper proposes Visual-Semantic Affect Graphs (VSAG), a graph-based framework that represents artworks through interpretable feature nodes, including color, brushstroke, composition, light-shadow, line quality, and extracted visual entities. The graph learns feature-emotion associations for dominant emotion prediction, with valence-arousal structure modeled through an optional hierarchical auxiliary affect module. We evaluate VSAG and VSAG-TopK against direct VLM inference using Qwen3-VL and InternVL3.5 on EmoArt-5k. Results show that direct VLMs provide useful but modest baselines, while graph-based models improve dominant emotion prediction. In particular, the hierarchical valence-arousal variant of VSAG achieves the best overall performance, improving accuracy, macro-F1, and weighted-F1 over both direct VLM inference and the emotion-only VSAG baseline. Per-class results indicate that graph-based models and VLMs capture partly complementary signals, while valence-arousal structure is most useful when integrated as a coarse-to-fine affect hierarchy rather than as a simple post-hoc correction. These findings suggest that explicit visual-semantic structure can support both prediction and interpretation in artwork emotion recognition.

\end{abstract}

\section{Introduction}
\label{sec:intro}

\begin{quote}
\itshape
``A work of art which did not begin in emotion is not art.''

\hfill --- Paul C\'ezanne ---
\end{quote}

This idea is not only an aesthetic statement, but also represents a potential computational challenge. If emotion is central to artistic meaning, then an artificial intelligence system that merely recognizes depicted objects, styles, or iconographic motifs has not yet understood an artwork in the way human viewers often do. Computationally, this requires connecting visual form to annotated affective response: how color, composition, texture, rhythm, light, abstraction, and symbolic association may contribute to how an artwork is labeled as feeling. In this sense, emotion prediction in art asks whether a model can move beyond visual description toward a deeper interpretation of aesthetic experience.

Vision-language models have shown strong performance in general visual-language understanding, suggesting that large-scale image-text pretraining can capture broad semantic knowledge from images~\cite{li2025survey}. However, affective understanding remains more difficult than recognizing objects, actions, or scenes. Human emotions are often implicit, context-dependent, subjective, and expressed through subtle visual or linguistic cues rather than directly observable categories. Recent studies suggest that VLMs can struggle with such affective reasoning, showing sensitivity to prompts and labels, bias toward frequent emotion categories, and difficulty distinguishing rare or complex emotions from more common neighboring ones~\cite{bhattacharyya2025evaluating,lei2024large,agarwal2026vision}. These limitations suggest that strong visual-language alignment alone may not be sufficient for robust emotion understanding.

This challenge becomes especially important in the study of artworks. Unlike many ordinary images, artworks evoke emotion not only through depicted objects or scenes, but also through formal and aesthetic components, like color, texture, composition, light, line, brushwork, spatial balance, and style. Psychological and neuroaesthetic accounts of art perception have long argued that aesthetic experience involves emotion, valuation, memory, and interpretation rather than visual recognition alone~\cite{pelowski2017move,mastandrea2019art}. Earlier computational studies on affective image and art analysis similarly emphasized that emotion prediction benefits from features motivated by psychology and art theory, including color, texture, composition, and semantic content~\cite{machajdik2010affective}. These observations suggest that progress in artwork emotion prediction may require not only larger models, but also intermediate representations that make affect-relevant visual and semantic cues explicit.

In this paper, we examine whether graph-based visual-semantic representations offer a more effective way to model emotion in art than direct VLM inference. Instead of treating an artwork as an isolated image, we represent it as an artwork-feature graph whose nodes encode affect-relevant attributes such as color, brushstroke, composition, light, line quality, and visual entities. The graph supports dominant emotion prediction, with valence-arousal group structure modeled through an optional hierarchical auxiliary module. We address three research questions:

\vspace{-10pt}

\begin{quote}
\textbf{RQ1:} How well do current vision-language models predict the emotions evoked by artworks?

\textbf{RQ2:} Can graph-based visual-semantic representations improve artwork emotion prediction beyond direct VLM inference?

\textbf{RQ3:} Which visual-semantic attributes are most strongly associated with specific emotions?
\end{quote}


The rest of this paper is organized as follows: Section~\ref{sec:rw} reviews some related works. Section~\ref{sec:method} introduces the proposed VSAG framework, including graph construction, VSAG-TopK, hierarchical valence-arousal auxiliary modeling, and training. Section~\ref{sec:res} presents the dataset, emotion prediction results, and feature-emotion association analysis. Section~\ref{sec:discussion} discusses the main findings and limitations. Finally, Section~\ref{sec:conclu} concludes the paper and outlines future directions.



\vspace{-10pt}

\section{Related Work}
\label{sec:rw}

\vspace{-10pt}
Research on emotion in visual art draws from both categorical and dimensional theories of affect. Discrete emotion models~\cite{ekman1971universals} describe affect through labels such as joy, sadness, fear, or anger, while dimensional models~\cite{russell1980circumplex} represent emotion through properties such as valence and arousal. This distinction is important for art because artworks often evoke emotion indirectly, through visual form, symbolic meaning, personal association, and cultural knowledge. Art-specific datasets such as WikiArt Emotions~\cite{mohammad2018wikiart}, ArtEmis~\cite{achlioptas2021artemis}, and FeelingBlue~\cite{ananthram2023feelingblue} have enabled larger-scale study of evoked emotion in artworks. However, recent AI-and-art research has increasingly focused on generative tasks, especially text-to-image generation, image editing, and diffusion-based visual art creation~\cite{wang2025diffusion,zhang2023text}. While these works show rapid progress in producing or editing artistic images, emotion prediction in art remains comparatively less explored. This gap is important because art emotion recognition is not only a labeling task, but also an interpretive problem involving visual form, semantic association, and affective response.

Vision-language models have become strong foundations for multimodal understanding, including image-text retrieval, captioning, visual question answering, and multimodal reasoning~\cite{li2025survey,liu2024mmbench,ying2024mmt}. However, strong general visual-language performance does not guarantee reliable affective understanding. Emotion is subjective, context-dependent, and often weakly grounded in visible objects. Prior work suggests that VLM predictions can be sensitive to prompts and label formulations, and that generic visual-language representations are not always aligned with human affective judgments~\cite{bondielli2021leveraging,bhattacharyya2025evaluating}. This limitation is especially relevant for art, where emotional meaning may depend on relations among color, form, composition, and symbolic content.

Graph representations provide a way to make visual and semantic structure explicit. In computer vision, scene graphs and knowledge graphs organize objects, attributes, relations, and contextual knowledge for visual reasoning~\cite{krishna2017visual}. Similar ideas have been used in computational art analysis, where resources such as ArtGraph~\cite{castellano2022leveraging} and ArtKB~\cite{blanco2026artkb} show that artwork understanding can benefit from structured information beyond pixels. Graph learning methods further provide tools for using relational structure in prediction~\cite{wu2020comprehensive,murgod2024survey}. However, it remains unclear whether graph-based learning over visual-semantic artwork attributes improves emotion prediction. Our work addresses this question by constructing an affect-focused artwork-feature graph and comparing it with direct VLM inference.

\section{Method}
\label{sec:method}

\begin{figure}[ht]
\centering
\includegraphics[width=\linewidth,keepaspectratio]{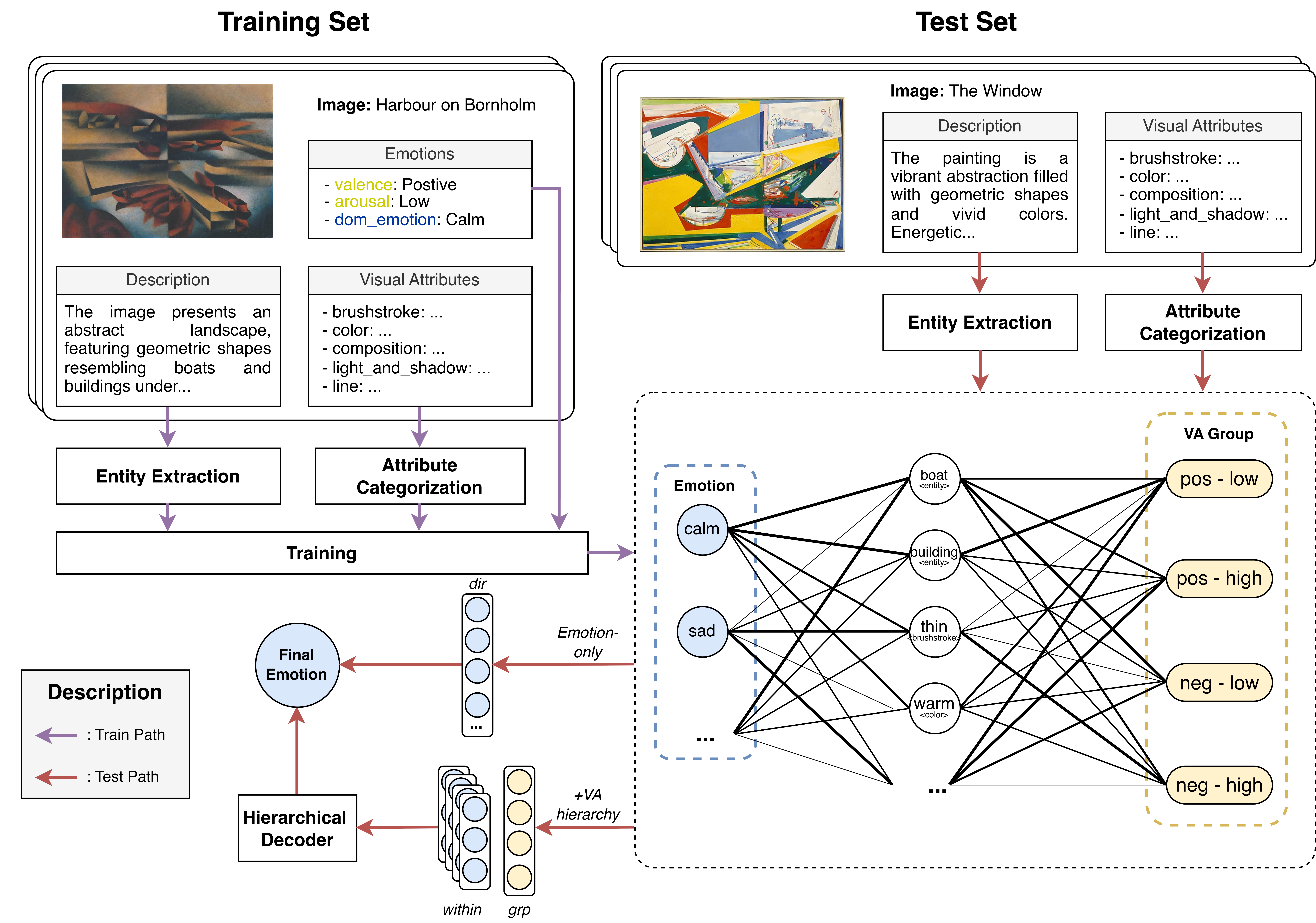}
\caption{Overview of the proposed Visual-Semantic Affect Graph framework.}
\label{fig:pipeline}
\end{figure}

\vspace{-20pt}
Figure~\ref{fig:pipeline} gives an overview of the proposed framework. We formulate artwork emotion prediction as structured prediction over a visual-semantic affect graph. Rather than directly mapping an image to an emotion label, we first represent each artwork through interpretable visual-semantic features and then learn how these features support dominant emotion prediction. The emotion-only model decodes the final emotion directly from graph-based emotion scores. The +VA variant uses the valence-arousal structure of the emotion label space as a coarse-to-fine hierarchy: the model predicts a four-way valence-arousal group and a within-group emotion distribution. In VSAG+VA, these outputs provide the hierarchical final decoder; in VSAG-TopK+VA, they are used as auxiliary regularization while final decoding remains direct.

\vspace{-10pt}

\subsection{Problem Setup and Graph Construction}

Let $\mathcal{D}=\{(x_i, y_i^{emo}, y_i^{grp})\}_{i=1}^{N}$
denote a dataset of artworks, where \(x_i\) is an artwork image, \(y_i^{emo}\) is its dominant emotion label, and \(y_i^{grp}\) is the valence-arousal group associated with that emotion. The main prediction target is \(y_i^{emo}\). Following the circumplex view of affect~\cite{russell1980circumplex}, \(y_i^{grp}\) is used to define the optional hierarchical auxiliary variant. 
For each artwork, we construct a visual-semantic feature representation from two natural-language sources: visual attribute descriptions and the general image description.




\subsubsection{Attribute Categorization}

Artwork images can often be described through basic visual elements such as color, line, texture, composition, and light. These elements provide a structured description of how the image is visually constructed, and they are important cues for affective interpretation. We first convert visual attribute descriptions into controlled categorical feature nodes. Specifically, we categorize brushstroke, color, composition, light-shadow, and line-quality descriptions using predefined lexical rules. This reduces noisy free-text descriptions into a smaller and more consistent feature vocabulary while preserving interpretable visual cues. Each matched category becomes a feature node connected to the corresponding artwork node.

\begin{table}[t]
\centering
\caption{Rule-based visual attribute categories used to construct feature nodes.}
\label{tab:visual_attribute_categories}
\scriptsize
\setlength{\tabcolsep}{3pt}
\renewcommand{\arraystretch}{1.15}
\begin{tabular}{p{0.14\linewidth}|p{0.8\linewidth}}
\toprule
\textbf{Attribute} & \textbf{Controlled category groups} \\
\midrule
Brushstroke &
Fine / Delicate; Detailed / Intricate; Smooth / Blended; Soft / Gentle; Rough / Textured; Expressive / Energetic; Movement / Flowing; Bold / Thick. \\
\midrule
Color &
Monochrome / Grayscale; Limited / Minimal Palette; High Contrast; Warm-Cool Contrast; Warm Colors; Cool Colors; High / Low Saturation; Vibrant / Intense; Muted / Subdued; Pastel; Dark / Bright Palette; Natural / Earthy; Complementary Colors; Harmonious / Balanced. \\
\midrule
Composition &
Balanced; Symmetrical; Asymmetrical; Central Focus; Multiple Focal Points; Dynamic; Flowing; Layered / Depth; Open Composition; Crowded / Dense; Scattered; Structured / Geometric; Minimalist; Chaotic / Tense; Intimate / Close. \\
\midrule
Light/shadow &
No Light Source / Abstract Lighting; Flat / Reduced Depth; Soft / Diffused Lighting; Harsh / Dramatic Lighting; Directional Lighting; Central Light Source; Ambient / Indirect Lighting; Strong / Subtle Contrast; Deep / Minimal / Soft / Defined Shadows; Illuminated / Highlighted; Adds Depth / Dimension; Soft Transitions. \\
\midrule
Line &
Blurred / Indistinct; Defined / Distinct; Thin / Fine; Thick / Bold; Delicate / Subtle; Smooth; Rough / Jagged; Firm / Decisive; Continuous / Broken; Flowing / Fluid; Curved / Organic; Precise / Controlled; Expressive / Energetic; Angular / Geometric; Structured; Elegant / Refined. \\
\bottomrule
\end{tabular}
\end{table}

The implementation also includes fallback categories such as \textit{Other} and \textit{Unknown / Not Applicable} when no rule matches or when the attribute is absent.

\vspace{-15pt}

\subsubsection{Entity Extraction}

Beyond formal attributes, an artwork may contain depicted entities or visual resemblances that provide semantic cues, especially in abstract art. These cues may include objects, motifs, scene elements, or concrete resemblances such as \textit{figure}, \textit{boat}, \textit{tree}, or \textit{circle}. Because such terms are open-vocabulary, we use a text LLM to extract depicted things from the artwork description while excluding formal visual-analysis terms. The prompt is abbreviated below:

\begin{lstlisting}[style=promptstyle]
Task: Extract only concrete depicted things, objects, motifs,
scene elements, or visual resemblances from the artwork description.

Keep examples: dot, circle, house, figure, face, tree, flower,
animal, bird, landscape, mosaic, geometric motif.

Exclude: color/palette, brushstroke/texture, line/edge,
composition/arrangement, lighting/shadow/depth, emotion/mood,
movement/rhythm, and generic visual words.

Rules: return lowercase singular noun phrases; remove color/style
adjectives unless essential; return [] if no concrete entity exists.

Output JSON only:
{"entities": ["entity one", "entity two"]}
\end{lstlisting}

We also apply lightweight post-processing to parse the JSON output, remove generic or formal-analysis terms, and deduplicate repeated entities.

\vspace{-15pt}

\subsubsection{Graph Definition}

We define a graph $G=(V,E)$, where the node set consists of artwork nodes, feature nodes, and affect label nodes $V = V_A \cup V_F \cup V_Y$. Here, $V_A$ is the set of artwork nodes, $V_F$ is the set of visual-semantic feature nodes, and $V_Y$ contains affect label nodes. In the emotion-only model, these are dominant emotion nodes. In the hierarchical auxiliary variant, \(V_Y\) also includes valence-arousal group labels and within-group emotion labels.

The observed graph contains artwork-feature edges $(a_i, f_j) \in E_{AF}$, where an edge exists if feature $f_j$ is observed in artwork $a_i$. The feature neighborhood of artwork $a_i$ is $\mathcal{N}(a_i)=\{f_j : (a_i,f_j)\in E_{AF}\}$. The model learns feature-label associations:
\[
f_j \rightarrow y_c^{(t)},
\]
where \(t\) denotes a prediction head, such as the direct emotion head, the valence-arousal group head, or the within-group emotion head, and \(y_c^{(t)}\) is a candidate label for that head. Therefore, each prediction can be decomposed into explicit graph paths:
\[
a_i \rightarrow f_j \rightarrow y_c^{(t)}.
\] and $y_c^{(t)}$ is a candidate label for that task. 


\vspace{-15pt}

\subsection{Visual-Semantic Affect Graph}
\vspace{-5pt}

We propose a Visual-Semantic Affect Graph (VSAG) model to predict affect labels by aggregating paths from an artwork node through its observed feature nodes to affect label nodes. To initialize feature-label associations, we estimate a smoothed log-odds score from the training set. For each feature $f_j$ and label $y_c^{(t)}$, we compute
\[
s_{j,c}^{(t)}
=
\log
\frac{
P(y_c^{(t)} \mid f_j)
}{
P(y_c^{(t)} \mid \neg f_j)
}.
\]

This score provides an initial estimate of the feature's association with an affect label. For example, if \textit{cool color} frequently appears in artworks labeled as \textit{Calm}, its initial feature-label score for \textit{Calm} is positive. These scores are then treated as learnable parameters and updated during training.

For each artwork $a_i$, candidate label $y_c^{(t)}$, and observed feature $f_j\in\mathcal{N}(a_i)$, VSAG computes a label-conditioned attention weight. Let $\mathbf{h}_j$ denote the representation of feature node $f_j$, constructed from its feature embedding and relation-type embedding. Let $\mathbf{u}_c^{(t)}$ denote the embedding of label $y_c^{(t)}$. The compatibility between feature $f_j$ and label $y_c^{(t)}$ is computed as
\[
g_{j,c}^{(t)}
=
\mathbf{w}^{\top}
\tanh(\mathbf{h}_j+\mathbf{u}_c^{(t)})
+
s_{j,c}^{(t)}.
\]

The label-conditioned attention weight is then
\[
\alpha_{i,j,c}^{(t)}
=
\frac{
\exp(g_{j,c}^{(t)})
}{
\sum_{f_k\in\mathcal{N}(a_i)}
\exp(g_{k,c}^{(t)})
}.
\]

The graph score for label $y_c^{(t)}$ is computed by aggregating artwork-feature-label paths:
\[
z_{i,c}^{(t)}
=
\sum_{f_j\in\mathcal{N}(a_i)}
\alpha_{i,j,c}^{(t)} s_{j,c}^{(t)}
+
b_c^{(t)},
\]
where $b_c^{(t)}$ is a task-specific label bias.

A small residual adjacency encoder can optionally be added:
\[
\tilde{z}_{i,c}^{(t)}
=
z_{i,c}^{(t)}
+
\lambda r_{i,c}^{(t)},
\]
where $r_{i,c}^{(t)}$ is produced from the artwork-feature adjacency vector and $\lambda$ controls the residual strength. The residual component is used only as a correction, while the primary prediction mechanism remains the explicit graph path score.

VSAG therefore assumes that affective meaning is distributed across the observed visual-semantic features. Each feature can contribute to each candidate affect label, with its contribution determined by both its feature-label edge score and its label-conditioned attention weight.

\vspace{-10pt}

\subsection{VSAG-TopK}

\vspace{-5pt}
We also test a variant called VSAG-TopK. It uses the same visual-semantic affect graph as VSAG, but changes how evidence is aggregated. While VSAG aggregates over all observed feature paths, VSAG-TopK assumes that an artwork's emotion may be determined by a small number of highly informative visual-semantic cues.

For each path $a_i\rightarrow f_j\rightarrow y_c^{(t)}$, VSAG-TopK computes a path score:
\[
p_{i,j,c}^{(t)}
=
s_{j,c}^{(t)}
+
\delta q_{j,c}^{(t)},
\]
where $s_{j,c}^{(t)}$ is the learned feature-label edge score, $q_{j,c}^{(t)}$ is a contextual compatibility score between feature $f_j$ and label $y_c^{(t)}$, and $\delta$ controls the strength of the contextual component.

Instead of computing an attention-weighted sum over all observed feature paths, VSAG-TopK selects the strongest $k$ paths for each candidate label. Let $\operatorname{TopK}(\mathcal{N}(a_i), c)$ denote the set of $k$ observed features with the largest path scores for label $y_c^{(t)}$. The label score is then
\[
z_{i,c}^{(t)}
=
\frac{1}{k}
\sum_{f_j\in \operatorname{TopK}(\mathcal{N}(a_i),c)}
p_{i,j,c}^{(t)}
+
b_c^{(t)}.
\]

Thus, VSAG and VSAG-TopK share the same graph structure but use different aggregation assumptions. VSAG treats affective meaning as distributed across all observed visual-semantic paths, while VSAG-TopK emphasizes the strongest feature-label paths.

\subsection{Hierarchical Valence-Arousal Auxiliary Modeling}

\vspace{-5pt}

In addition to the emotion-only model, we evaluate a hierarchical valence-arousal auxiliary variant. The motivation is that the emotion label space has a known coarse-to-fine structure: each dominant emotion belongs to a valence-arousal group defined by the dataset label scheme. Let \(\mathcal{Q}\) denote the set of valence-arousal groups, and let \(q(e)\in\mathcal{Q}\) denote the group associated with emotion \(e\). The full emotion-to-group mapping is given in Section~\ref{subsec:ds}.

The model computes three graph-based outputs. First, it computes a direct emotion distribution
\[
p_i^{dir}(e)=\mathrm{softmax}(z_i^{emo})_e .
\]
Second, it computes a valence-arousal group distribution
\[
p_i^{grp}(q)=\mathrm{softmax}(z_i^{grp})_q .
\]
Third, for each valence-arousal group, it computes a within-group emotion distribution over the emotions belonging to that group:
\[
p_i^{within}(e \mid q(e)) =
\frac{\exp(z_i^{within}(e))}
{\sum_{e' \in \mathcal{E}_{q(e)}} \exp(z_i^{within}(e'))},
\]
where \(\mathcal{E}_{q}\) is the set of emotions in group \(q\).

The hierarchical emotion distribution is then defined as
\[
p_i^{hier}(e)=p_i^{grp}(q(e)) \, p_i^{within}(e \mid q(e)).
\]

Unlike a simple additive valence-arousal decoder, this formulation uses valence-arousal structure as a coarse-to-fine hierarchy: the model first estimates the affective group and then selects an emotion within that group. For example, a Positive-Low prediction narrows the space to emotions such as \textit{Calm}, \textit{Contentment}, and \textit{Glad}.

For the emotion-only variant, the final prediction is obtained from \(p_i^{dir}\). For the reported VSAG+VA variant, the final prediction is obtained from the hierarchical distribution \(p_i^{hier}\). For VSAG-TopK+VA, the valence-arousal hierarchy is used only as an auxiliary regularizer during training, while final decoding remains direct.

\vspace{-15pt}

\subsection{Training and Inference}
\vspace{-5pt}

The emotion-only model is trained with the dominant emotion loss: $\mathcal{L}=\mathcal{L}_{emo}.$ The hierarchical +VA variant adds auxiliary losses for the valence-arousal group and within-group emotion prediction:
\vspace{-5pt}
\[
\mathcal{L}
=
\lambda_{emo}\mathcal{L}_{emo}
+
\lambda_{grp}\mathcal{L}_{grp}
+
\lambda_{within}\mathcal{L}_{within}
+
\lambda_{cons}\mathcal{L}_{cons}.
\]
Here, \(\mathcal{L}_{grp}\) supervises the four-way valence-arousal group prediction, and \(\mathcal{L}_{within}\) supervises emotion prediction within the gold group. The consistency term \(\mathcal{L}_{cons}\) encourages the predicted group distribution to agree with the group distribution implied by the emotion probabilities. Setting \(\lambda_{grp}=\lambda_{within}=\lambda_{cons}=0\) recovers the emotion-only model.

During training, the model learns feature embeddings, relation embeddings, affect label embeddings, and feature-label edge scores. For the hierarchical auxiliary variant, the model jointly optimizes the direct emotion head, the valence-arousal group head, and the within-group emotion head. Thus, valence-arousal information is not used as a post-hoc additive logit correction; instead, it provides a coarse-to-fine affect structure for regularization or hierarchical decoding.

At inference time, the emotion-only model predicts the emotion with the highest direct graph-based emotion probability. The hierarchical +VA variant additionally predicts a valence-arousal group and a within-group emotion distribution. For VSAG+VA, the final emotion is decoded from the hierarchical distribution \(p_i^{hier}(e)\). For VSAG-TopK+VA, the auxiliary valence-arousal losses are used during training, while the final prediction is decoded from the direct emotion head.

Because prediction is based on explicit graph paths, the model can provide path-level explanations. For VSAG, the contribution of feature $f_j$ to label $y_c^{(t)}$ is $\alpha_{i,j,c}^{(t)} s_{j,c}^{(t)}$. For VSAG-TopK, the explanation consists of the highest-scoring feature-label paths selected by the TopK pooling operation. These explanations identify which visual-semantic attributes most strongly support a predicted emotion or valence-arousal group label.

\vspace{-15pt}

\section{Results}
\label{sec:res}

\vspace{-10pt}

\subsection{Dataset}
\label{subsec:ds}
\vspace{-5pt}

We use EmoArt-5k, a publicly available subset\footnote{\url{https://huggingface.co/datasets/printblue/EmoArt-5k}} of the EmoArt dataset~\cite{zhang2025emoart}. The dataset contains abstract artwork images paired with natural-language descriptions, visual attribute annotations, and affective labels. The dominant emotion labels follow the valence-arousal grouping used by EmoArt:
\[
\begin{aligned}
(\text{Positive}, \text{Low}) &: 
\{\text{Calm (\texttt{Cal})}, \text{Contentment (\texttt{Con})}, \text{Glad (\texttt{Gla})}\}, \\
(\text{Positive}, \text{High}) &: 
\{\text{Excited (\texttt{Exc})}, \text{Happy (\texttt{Hap})}, \text{Aroused (\texttt{Aro})}\},\\
(\text{Negative}, \text{Low}) &: 
\{\text{Sad (\texttt{Sad})}, \text{Bored (\texttt{Bor})}, \text{Tired (\texttt{Tir})}\}, \\
(\text{Negative}, \text{High}) &: 
\{\text{Alarmed (\texttt{Alm})}, \text{Frustrated (\texttt{Fru})}, \text{Annoyed (\texttt{Ann})}\}.
\end{aligned}
\]

Here, we use the dominant emotion label as the main prediction target, while the corresponding valence-arousal groups are used only in auxiliary variants.


\begin{table}[ht]
\centering
\caption{Distribution of dominant emotion labels by valence-arousal group.}
\label{tab:emotion_distribution}
\scriptsize
\setlength{\tabcolsep}{4pt}
\renewcommand{\arraystretch}{1.05}
\begin{tabular}{l|l|r||l|l|r}
\toprule
\textbf{Emotion} & \textbf{VA group} & \textbf{Count} &
\textbf{Emotion} & \textbf{VA group} & \textbf{Count} \\
\midrule
Calm        & Positive-Low  & 2872 & Sad        & Negative-Low  & 216 \\
Contentment & Positive-Low  & 692  & Bored      & Negative-Low  & 25  \\
Glad        & Positive-Low  & 2    & Tired      & Negative-Low  & 19  \\
\midrule
Excited     & Positive-High & 1240 & Alarmed    & Negative-High & 207 \\
Happy       & Positive-High & 50   & Frustrated & Negative-High & 157 \\
Aroused     & Positive-High & 24   & Annoyed    & Negative-High & 28  \\
\bottomrule
\end{tabular}
\end{table}

The label distribution is highly imbalanced, as shown in Table~\ref{tab:emotion_distribution}. The dataset is dominated by the \textit{Calm} class, which accounts for more than half of all artworks. In contrast, several emotion classes contain only a small number of examples, such as \textit{Glad}, \textit{Aroused}, \textit{Bored}, \textit{Tired}, and \textit{Annoyed}. This imbalance makes the task challenging because a model can obtain high overall accuracy by favoring frequent emotions, while still performing poorly on rare affective categories. For this reason, we report both distribution-sensitive metrics, such as accuracy and weighted-F1, and class-balanced metrics, such as macro-F1.

After applying the visual attribute categorization and entity extraction steps described in Section~\ref{sec:method}, we obtain the visual-semantic feature statistics summarized in Table~\ref{tab:dataset_feature_statistics}. Each artwork contains a short description of about 36 words on average. The visual attribute fields produce compact feature sets, with roughly 2 to 3 categorical nodes per attribute type. The entity extractor adds a more open-ended semantic layer, contributing 4.06 normalized entity nodes per artwork on average, and 5,281 unique normalized visual entity nodes over the dataset.

\vspace{-15pt}

\begin{table}[ht]
\centering
\caption{Dataset feature statistics. For description length and feature fields, values are reported as mean, minimum, and maximum per artwork.}
\label{tab:dataset_feature_statistics}
\scriptsize
\setlength{\tabcolsep}{3.5pt}
\renewcommand{\arraystretch}{1.08}
\begin{tabular}{l|rrr|rrr|rrr}
\toprule
\multirow{2}{*}{\textbf{Field}} 
& \multicolumn{3}{c}{\textbf{Train}} 
& \multicolumn{3}{c}{\textbf{Test}} 
& \multicolumn{3}{c}{\textbf{All}} \\
\cmidrule(lr){2-4}
\cmidrule(lr){5-7}
\cmidrule(lr){8-10}
& \textbf{Mean} & \textbf{Min} & \textbf{Max}
& \textbf{Mean} & \textbf{Min} & \textbf{Max}
& \textbf{Mean} & \textbf{Min} & \textbf{Max} \\
\midrule
Description words      & 35.73 & 24 & 53 & 35.66 & 28 & 56 & 35.72 & 24 & 56 \\
Brushstroke categories & 1.97  & 1  & 5  & 1.97  & 1  & 4  & 1.97  & 1  & 5  \\
Color categories       & 2.87  & 1  & 7  & 2.87  & 1  & 7  & 2.87  & 1  & 7  \\
Composition categories & 2.01  & 1  & 5  & 1.97  & 1  & 5  & 2.00  & 1  & 5  \\
Light-shadow categories& 2.17  & 1  & 7  & 2.20  & 1  & 6  & 2.18  & 1  & 7  \\
Line categories& 2.91  & 1  & 7  & 2.95  & 1  & 6  & 2.92  & 1  & 7  \\
Visual entities        & 4.07  & 0  & 15 & 4.00  & 0  & 11 & 4.06  & 0  & 15 \\
\midrule
Artworks               & \multicolumn{3}{c}{4426} & \multicolumn{3}{c}{1106} & \multicolumn{3}{c}{5532} \\
Unique visual entities & \multicolumn{3}{c}{4547} & \multicolumn{3}{c}{1744} & \multicolumn{3}{c}{5281} \\
\bottomrule
\end{tabular}
\end{table}

\vspace{-35pt}

\subsection{Emotion Prediction Performance}
\label{subsec:emotion_prediction}

\vspace{-5pt}

\begin{table}[t]
\centering
\caption{Dominant emotion prediction performance. Best values are shown in bold.}
\label{tab:emotion_prediction_overall}
\scriptsize
\setlength{\tabcolsep}{3pt}
\begin{tabular}{l|ccc}
\toprule
\textbf{Model} & \textbf{Acc.} & \textbf{Macro-F1} & \textbf{Weighted-F1} \\
\midrule
Qwen3-VL-8B    & 0.480 & 0.168 & 0.402 \\
Qwen3-VL-4B    & 0.507 & 0.158 & 0.433 \\
Qwen3-VL-2B    & 0.505 & 0.159 & 0.430 \\
InternVL3.5-8B & 0.549 & 0.206 & 0.536 \\
InternVL3.5-4B & 0.403 & 0.136 & 0.413 \\
InternVL3.5-2B & 0.353 & 0.130 & 0.374 \\
\midrule
\textbf{VSAG}        & 0.577 & 0.187 & 0.571 \\
\textbf{VSAG+VA}     & \textbf{0.591} & \textbf{0.214} & \textbf{0.580} \\
\textbf{VSAG-TopK}   & 0.579 & 0.201 & 0.569 \\
\textbf{VSAG-TopK+VA} & 0.554 & 0.193 & 0.552 \\
\bottomrule
\end{tabular}
\end{table}

\begin{table}[t]
\centering
\caption{Per-class F1 scores for dominant emotion prediction. Best values are shown in bold.}
\label{tab:emotion_prediction_per_class}
\scriptsize
\setlength{\tabcolsep}{1pt}
\renewcommand{\arraystretch}{1.04}
\begin{tabular}{l|cccccccccccc}
\toprule
\textbf{Model}
& \textbf{\texttt{Cal}} & \textbf{\texttt{Con}} & \textbf{\texttt{Gla}}
& \textbf{\texttt{Exc}} & \textbf{\texttt{Hap}} & \textbf{\texttt{Aro}}
& \textbf{\texttt{Sad}} & \textbf{\texttt{Bor}} & \textbf{\texttt{Tir}}
& \textbf{\texttt{Alm}} & \textbf{\texttt{Fru}} & \textbf{\texttt{Ann}} \\
\midrule
Qwen3-VL-8B 
& \textbf{0.735} & 0.000 & 0.000
& 0.015 & \textbf{0.353} & 0.024
& 0.339 & \textbf{0.133} & 0.067
& 0.290 & 0.056 & 0.000 \\

Qwen3-VL-4B 
& \textbf{0.735} & 0.000 & 0.000
& 0.148 & 0.145 & 0.030
& \textbf{0.376} & 0.100 & 0.000
& 0.359 & 0.000 & 0.000 \\

Qwen3-VL-2B 
& 0.701 & 0.000 & 0.000
& 0.188 & 0.110 & 0.000
& 0.360 & 0.000 & 0.000
& \textbf{0.459} & 0.089 & 0.000 \\

InternVL3.5-8B 
& 0.713 & 0.000 & 0.000
& \textbf{0.626} & 0.263 & \textbf{0.091}
& 0.287 & 0.000 & \textbf{0.071}
& 0.371 & 0.051 & 0.000 \\

InternVL3.5-4B 
& 0.646 & 0.000 & 0.000
& 0.299 & 0.250 & 0.053
& 0.192 & 0.000 & 0.000
& 0.190 & 0.000 & 0.000 \\

InternVL3.5-2B 
& 0.552 & 0.000 & 0.000
& 0.320 & 0.124 & 0.027
& 0.189 & 0.000 & 0.000
& 0.297 & 0.049 & 0.000 \\

\midrule
\textbf{VSAG}
& 0.734 & \textbf{0.280} & 0.000
& 0.624 & 0.095 & 0.000
& 0.158 & 0.000 & 0.000
& 0.232 & 0.125 & 0.000 \\

\textbf{VSAG+VA}
& 0.733 & 0.272 & 0.000
& 0.624 & 0.118 & 0.000
& 0.247 & 0.000 & 0.000
& 0.317 & \textbf{0.254} & 0.000 \\

\textbf{VSAG-TopK}
& 0.734 & 0.252 & 0.000
& 0.607 & 0.182 & 0.000
& 0.164 & 0.000 & 0.000
& 0.361 & 0.118 & 0.000 \\

\textbf{VSAG-TopK+VA}
& 0.706 & 0.259 & 0.000
& 0.596 & 0.087 & 0.000
& 0.143 & 0.000 & 0.000
& 0.286 & 0.239 & 0.000 \\
\bottomrule
\end{tabular}
\end{table}



We evaluate direct vision-language model inference using Qwen3-VL and InternVL3.5 at multiple model sizes. For these baselines, the model is given the artwork image and prompted to select the dominant emotion label directly. This setting evaluates whether current VLMs can infer evoked emotion from the image without the explicit visual-semantic graph structure used by our method.

Table~\ref{tab:emotion_prediction_overall} reports overall dominant emotion classification performance. For \textbf{RQ1}, direct VLM inference provides a useful but still modest baseline. The best direct VLM is InternVL3.5-8B, with accuracy and weighted-F1 only slightly above 0.5. This result suggests that current VLMs can capture some broad affective cues, but artwork emotion prediction remains difficult because the emotional meaning of art is often implicit, subjective, and grounded in subtle relations among form, composition, and semantic association. This difficulty is not surprising, since interpreting emotion in art is challenging even for human viewers and often depends on context and personal perception. 

The VLM results also show that larger parameter size does not necessarily lead to better emotion prediction. For example, Qwen3-VL-8B performs worse than its 2B and 4B variants on overall accuracy. One possible explanation is that direct emotion prediction is sensitive to prompt interpretation, label priors, and the model's alignment behavior, rather than only to raw model capacity. Larger models may produce more complex or language-prior-driven judgments, but those judgments do not always match the true emotion.

For \textbf{RQ2}, the graph-based models generally outperform direct VLM inference on accuracy and weighted-F1. The strongest overall model is \textbf{VSAG+VA}, which achieves the best score across all three metrics. Compared with both the best direct VLM and the emotion-only VSAG baseline, this result shows that hierarchical valence-arousal modeling can improve dominant emotion prediction when it is integrated as a coarse-to-fine affect structure rather than as a post-hoc logit correction. The effect of VA is not uniform across graph aggregation strategies. While VSAG+VA improves over VSAG, VSAG-TopK+VA underperforms VSAG-TopK. This suggests that the VA hierarchy is more useful when affective evidence is aggregated across distributed feature paths, whereas sparse TopK pooling may already impose a strong evidence-selection bias.

Table~\ref{tab:emotion_prediction_per_class} shows that the gains are concentrated in several negative or difficult classes. Compared with emotion-only VSAG, VSAG+VA improves \textit{Alarmed}, \textit{Frustrated}, and \textit{Sad}, which contributes to its higher macro-F1. However, rare classes such as \textit{Glad}, \textit{Aroused}, \textit{Bored}, \textit{Tired}, and \textit{Annoyed} remain difficult. Direct VLMs also remain stronger for some classes, suggesting that VLMs and graph-based models capture partly complementary affective signals.

\vspace{-10pt}

\subsection{Feature-Emotion Association Analysis}
\label{subsec:feature_emotion_analysis}
\vspace{-5pt}
To answer \textbf{RQ3}, we inspect the learned associations between visual-semantic feature nodes and emotion labels. Table~\ref{tab:feature_emotion_associations} reports the top associated feature nodes for each emotion. The associations are obtained from feature-emotion log-odds scores estimated from the training set and used to initialize the VSAG feature-emotion edges. Each score indicates how strongly a feature is associated with an emotion label relative to its absence in the training set.

The associations reveal several interpretable patterns. High-arousal or negative emotions are often linked to visually intense features: \textit{Alarmed} is associated with strong contrast, tense composition, and dramatic lighting, while \textit{Frustrated} is associated with rough or energetic brushwork and strong contrast. In contrast, calmer positive emotions are linked to softer visual properties: \textit{Calm} is associated with soft lighting, gentle brushwork, and balanced composition, while \textit{Contentment} is associated with warm colors and domestic entities such as \textit{baby}.

\begin{table}[htb]
\centering
\caption{Top-3 visual-semantic feature associations for each emotion. Each entry shows the feature node type, feature value, and association score in parentheses.}
\label{tab:feature_emotion_associations}
\tiny
\setlength{\tabcolsep}{2pt}
\renewcommand{\arraystretch}{1.12}
\begin{tabular}{p{0.155\linewidth}|p{0.155\linewidth}|p{0.155\linewidth}|p{0.155\linewidth}|p{0.155\linewidth}|p{0.155\linewidth}}
\toprule
\textbf{Calm} & \textbf{Contentment} & \textbf{Glad} & \textbf{Excited} & \textbf{Happy} & \textbf{Aroused} \\
\midrule
Light/shadow: Soft / Diffused Lighting (0.49) &
Entity: baby (0.99) &
Entity: forest (4.99) &
Color: Vibrant / Intense (1.62) &
Entity: blue sky (3.05) &
Entity: vertical element (3.37) \\

Brushstroke: Soft / Gentle (0.44) &
Entity: sofa (0.93) &
Entity: zigzag (4.87) &
Color: High Saturation (1.51) &
Entity: cat (2.92) &
Entity: linear element (3.37) \\

Composition: Balanced (0.43) &
Color: Warm Colors (0.84) &
Composition: Minimalist (4.87) &
Composition: Dynamic (1.33) &
Color: Warm Colors (2.12) &
Composition: Minimalist (2.62) \\

\midrule
\textbf{Sad} & \textbf{Bored} & \textbf{Tired} & \textbf{Alarmed} & \textbf{Frustrated} & \textbf{Annoyed} \\
\midrule
Color: Muted / Subdued (1.76) &
Entity: linear element (3.21) &
Entity: introspection (3.48) &
Light/shadow: Strong / High Contrast (2.45) &
Brushstroke: Rough / Textured (1.94) &
Entity: pedestrian (3.42) \\

Color: Dark Palette (1.74) &
Entity: waterfront (3.15) &
Entity: dark suit (3.48) &
Composition: Chaotic / Tense (2.26) &
Brushstroke: Expressive / Energetic (1.89) &
Entity: facial feature (3.42) \\

Entity: soldier (1.65) &
Composition: Minimalist (2.47) &
Line: Blurred / Indistinct Lines (2.88) &
Light/shadow: Harsh / Dramatic Lighting (2.12) &
Light/shadow: Strong / High Contrast (1.86) &
Composition: Minimalist (2.67) \\
\bottomrule
\end{tabular}
\end{table}

\begin{figure}[htb]
\centering
\includegraphics[width=.75\linewidth,keepaspectratio]{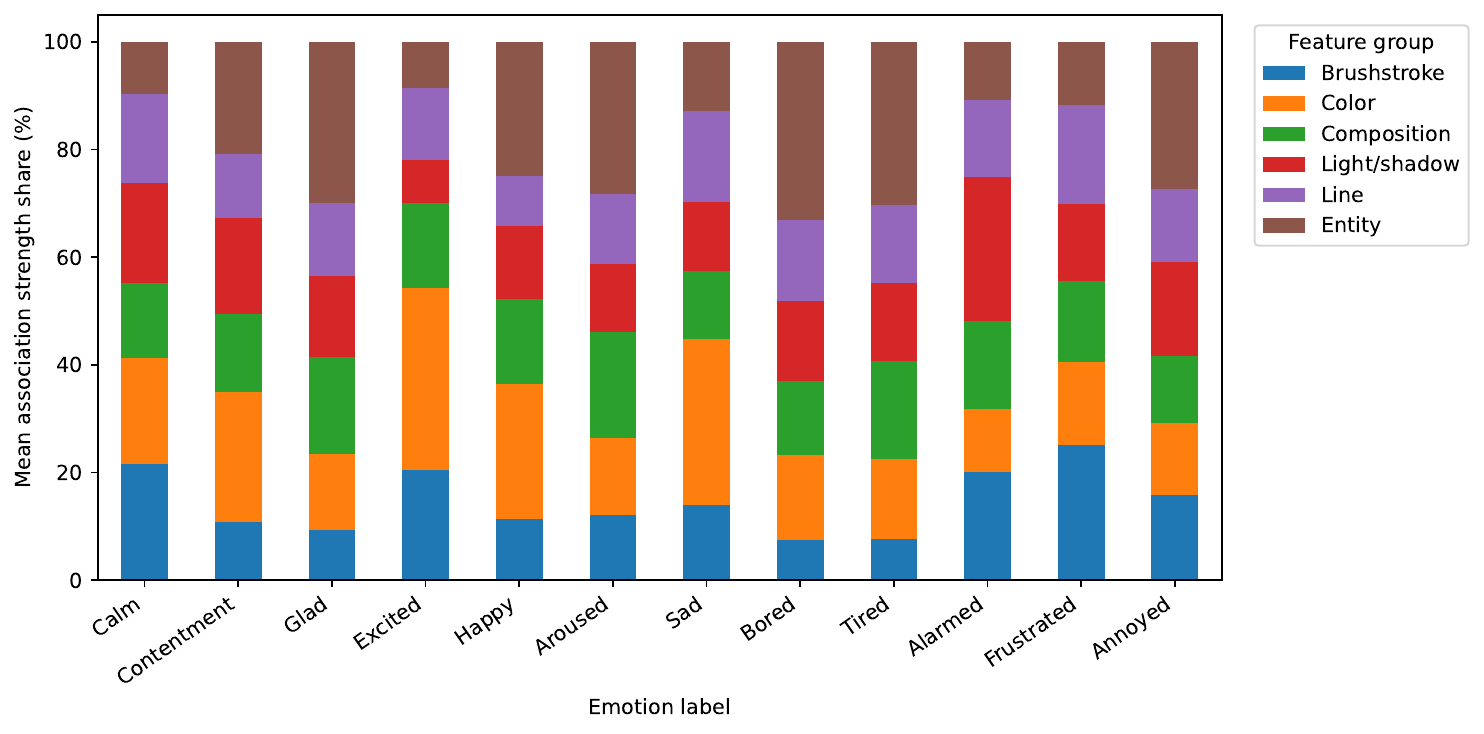}
\caption{Feature-group mean association strength by emotion.}
\label{fig:feature_group_association_strength}
\end{figure}

\vspace{-20pt}

Figure~\ref{fig:feature_group_association_strength} summarizes the same associations at the feature-group level. It uses all positive feature-emotion associations whose feature node appears in at least five training artworks, computes the mean association strength for each feature group, and normalizes these values within each emotion. The results indicate a relatively balanced contribution across feature groups, with no single group consistently dominating across all emotion labels: color is prominent for \textit{Excited} and \textit{Sad}, light-shadow for \textit{Alarmed}, brushstroke for \textit{Frustrated} and \textit{Calm}, and entity nodes for labels such as \textit{Glad}, \textit{Bored}, \textit{Tired}, and \textit{Annoyed}.

However, note that these values should be interpreted as learned feature-emotion associations rather than causal feature importance. Some associations, especially for rare classes such as \textit{Glad} may reflect limited support and should therefore be interpreted cautiously. Accordingly, this analysis is best interpreted as descriptive evidence of which visual-semantic feature groups are most strongly associated with each emotion label.

\vspace{-15pt}

\section{Discussion}
\label{sec:discussion}
\vspace{-10pt}

Our results show that explicit visual-semantic structure is useful for artwork emotion prediction, but the task remains difficult. One reason is that emotion in art is rarely limited to a single label: an artwork may evoke multiple emotions, and different viewers may interpret it differently. Since the dataset provides only one dominant emotion label per artwork, models are trained to match a specific annotation choice rather than the full range of possible affective responses. The graph formulation partly addresses this limitation by producing scores for all emotion nodes, which allows secondary emotions and feature-emotion paths to be inspected.

The comparison with direct VLM inference suggests that VLMs and graph-based models capture complementary signals. VLMs may rely on global visual impressions and broad semantic priors, while VSAG uses explicit feature nodes such as color, brushstroke, composition, light-shadow, line quality, and visual entities. This makes VSAG more interpretable, but also dependent on feature quality and coverage. The +VA results show that valence-arousal hierarchy improves standard VSAG, mainly by helping several negative or less frequent classes. However, this benefit does not transfer to VSAG-TopK, suggesting that VA structure is most useful with distributed feature-path aggregation, while TopK pooling may already emphasize dominant affective cues. The persistent near-zero performance on rare labels shows that long-tailed affect supervision remains a major challenge, motivating future investigation of class-reweighting, focal loss, or graph-based sampling strategies. Multi-label emotion annotation may also better reflect the range of affective responses that an artwork can evoke.

Finally, a further limitation is that the current graph mainly uses visual-semantic features and does not yet incorporate non-visual contextual information such as artist, year, medium, style period, or cultural background, which may affect emotional interpretation. Previous work has shown that such contextual knowledge can be represented through knowledge graphs and combined with visual information for artwork emotion recognition~\cite{castellano2023recognizing}. In addition, our current graph-based evaluation primarily compares variants of VSAG rather than independently developed graph-based approaches. Future work could therefore incorporate external art knowledge into VSAG to construct a richer artwork knowledge graph and evaluate it against existing knowledge-graph-based methods.

\vspace{-15pt}

\section{Conclusion}
\label{sec:conclu}
\vspace{-10pt}
Our work studied artwork emotion prediction using Visual-Semantic Affect Graphs, which represent artworks through interpretable visual and semantic feature nodes and learn their associations with emotion labels. Experiments on EmoArt-5k show that direct VLM inference provides a useful but modest baseline, while graph-based models improve dominant emotion prediction. The best overall model is VSAG+VA, which integrates valence-arousal structure through a hierarchical auxiliary affect formulation and achieves the highest accuracy, macro-F1, and weighted-F1 among the evaluated models. The results also show that the benefit of VA structure depends on the graph aggregation strategy: it improves standard VSAG, but not VSAG-TopK. Per-class analysis shows that hierarchical VA modeling improves several difficult negative emotions, while rare classes remain challenging because of severe label imbalance. Feature-emotion analysis identifies interpretable associations between emotions and visual-semantic cues such as contrast, brushwork, composition, lighting, and entities. Future work can extend VSAG toward multi-label emotion modeling, richer art-historical metadata, and hybrid models that combine VLM representations with interpretable graph structure.


%
%
%
\vspace{-15pt}
\bibliographystyle{splncs04}
\bibliography{mybibliography}
%




\end{document}